\documentclass[sigconf]{acmart}

\renewcommand\footnotetextcopyrightpermission[1]{} 
\def\BibTeX{{\rm B\kern-.05em{\sc i\kern-.025em b}\kern-.08emT\kern-.1667em\lower.7ex\hbox{E}\kern-.125emX}}
    
\usepackage{placeins}
\usepackage{hyperref}
\usepackage{subfig}
\usepackage{multirow}
\usepackage{tikz}
\usepackage{svg}
\usepackage{todonotes}
\usepackage{multicol}
\usepackage{mathrsfs}
\usepackage{flushend}
\usepackage[super]{nth}    
    
\setcopyright{none}
\begin{document}
%
\title{HyperLearn: A Distributed Approach for Representation Learning in Datasets With Many Modalities}

\author{Devanshu Arya}
\affiliation{%
  \institution{University of Amsterdam}
  \city{Amsterdam}
  \country{The Netherlands}}
\email{d.arya@uva.nl}

\author{Stevan Rudinac}
\affiliation{%
  \institution{University of Amsterdam}
  \city{Amsterdam}
  \country{The Netherlands}}
\email{s.rudinac@uva.nl}

\author{Marcel Worring}
\affiliation{%
  \institution{University of Amsterdam}
  \city{Amsterdam}
  \country{The Netherlands}}
\email{m.worring@uva.nl}

\renewcommand{\shorttitle}{HyperLearn: A Distributed Approach for Representation Learning}

\begin{abstract}
   Multimodal datasets contain an enormous amount of relational information, which grows exponentially with the introduction of new modalities.  Learning representations in such a scenario is inherently complex due to the presence of multiple heterogeneous information channels. These channels can encode both (a) inter-relations between the items of different modalities and (b) intra-relations between the items of the same modality. Encoding multimedia items into a continuous low-dimensional semantic space such that both types of relations are captured and preserved is extremely challenging, especially if the goal is a unified end-to-end learning framework. The two key challenges that need to be addressed are: 1) the framework must be able to merge complex intra and inter relations without losing any valuable information and 2) the learning model should be invariant to the addition of new and potentially very different modalities. In this paper, we propose a flexible framework which can scale to data streams from many modalities. To that end we introduce a hypergraph-based model for data representation and deploy Graph Convolutional Networks to fuse relational information within and across modalities. Our approach provides an efficient solution for distributing otherwise extremely computationally expensive or even unfeasible training processes across multiple-GPUs, without any sacrifices in accuracy. Moreover, adding new modalities to our model requires only an additional GPU unit keeping the computational time unchanged, which brings representation learning to truly multimodal datasets. We demonstrate the feasibility of our approach in the experiments on multimedia datasets featuring second, third and fourth order relations.
   

\end{abstract}



\maketitle

\section{Introduction}

\begin{figure*}[t]
 \centering
 \includegraphics[width=0.85\textwidth]{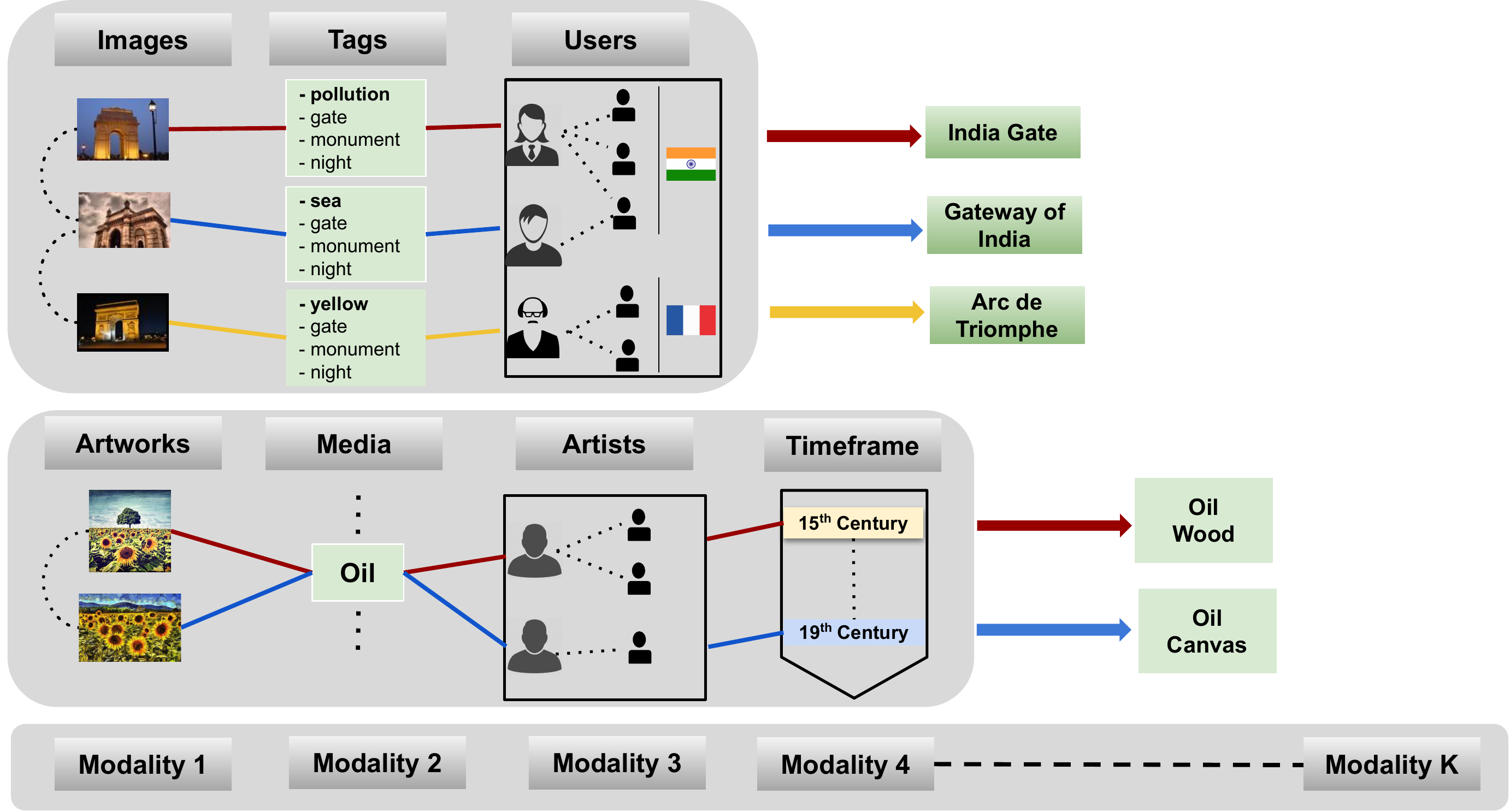}

 \centerline{}\medskip
\caption{Example showing importance of capturing ternary  relations (images-tags-users) in a social network dataset and quaternary relations (artworks-media-artists-timeframe) in artistic dataset. HyperLearn exploits such relations to learn complex representations for each modality. At the same time, HyperLearn provides a distributed learning approach, which makes it scalable to datasets with many modalities }
\label{intro_fig}
\end{figure*}

The field of multimedia has been slowly, but steadily growing beyond simple combining of diverse modalities, such as text, audio and video, to modeling their complex relations and interactions. 
These relations are commonly perceived, and therefore, modelled as only pair-wise connections between two items, which is a major drawback in the majority of the existing techniques. Going beyond pair-wise connections to encode higher-order relations can not only discover complex inter-dependencies between items but also help in removing ambiguous relations. For instance: in the task of social image-tag refinement, conventional approaches focus on exploiting the pairwise tag-image relations, without considering the user information which has been proven extremely useful in resolving tag ambiguities and closing the semantic gap between visual representation and semantic meaning \cite{cui2014social, tang2017tri, tang2019social, li2016weakly}. It is hence an interesting, but far more challenging problem in multimedia to exploit and learn higher-order relations to be able to (a) learn a better representation for each item, (b) improve pairwise retrieval tasks and (c) discover far more complex relations which can be ternary ($\nth{3}$ order), quaternary ($\nth{4}$ order), quinary ($\nth{5}$ order) or even beyond. As examples, figure~\ref{intro_fig} shows the importance of modeling higher-order relations in social networks and in artistic analysis respectively. In the upper example from Figure~\ref{intro_fig}, textual annotations and information about user demographics is utilized for disambiguation between landmarks with very similar visual appearance. Similarly, the second example illustrates quaternary relations formed by the artworks, media, artists and the time-frame in which they were active. Capturing such complex relations is of utmost importance in a number of tasks performed by the domain experts, such as author attribution, influence and appreciation analysis. 

Learning representations in multimodal datasets is an extremely complex task due to the enormous amount of relational information available. At the same time most of these relations have an innate property of `homophily', which is the fact that similarity breeds connections. Exploiting this property of similarities can help in immensely simplifying the understanding of these relations. These similarities can be derived from both intra relations between items of the same modality and inter-relations between items across different modalities. Unifying the two types of relations in a complementary manner has the potential to bolster the performance of practically any multimedia task. Thus, in this work we propose an efficient learning framework that can merge information generated by both intra as well as inter-relations in datasets with many modalities. We conjecture that such an approach can pave the way for a generic methodology for learning representations by exploiting higher-order relations. At the same time, we introduce an approach that makes our framework scale to multiple modalities.

We focus on learning a low-dimensional representation for each multimodal item using an unsupervised framework. The unsupervised methods utilize relational information both within as well as across modalities to learn common representations for a given multimodal dataset. The co-occurrence information simply means that two items from different modalities are semantically similar if they co-exist in a multimedia collection. For example, the textual description of a video often describes the events shown in the visual channel. Many of the multimedia tasks revolve around this compact latent representation of each multimodal entity ~\cite{rasiwasia2010new, ngiam2011multimodal}. The major challenge lies in bridging the learning gap between the two types of relations in a way that they can be semantically complementary in describing similar concepts. Learning representation is usually extremely expensive, both in computational time and required storage as even a relatively small multimedia collection normally contains a multitude of complex relations. 


Handling a large amount of relations requires a framework with a flexible approach to training across multiple pipelines. Most of the existing algorithms fail in parallelizing their framework into separate pipelines ~\cite{li2018deep, yang2015network}, resulting in large time and memory consumption. Thus, in the proposed framework we can parallelize the training process for different modalities into separate pipelines, each requiring just an additional GPU core. By doing so, we facilitate joint multimodal representation learning on highly heterogeneous multimedia collections containing an arbitrarily large number of modalities, effectively hitting an elusive target sought after since the early days of multimedia research. 
The points below highlight the contributions of this paper:
\begin{itemize}

\item We address the challenging problem of multimodal representation learning by proposing HyperLearn, an unsupervised framework capable of jointly modeling relations between the items of the same modality, as well as across different modalities. 

\item Based on the concept of geometric deep learning on hypergraphs, our HyperLearn framework is effective in extracting higher-order relations in multimodal datasets. 

\item In order to reduce prohibitively high computational costs associated with multimodal representation learning, in this work we propose a distributed learning approach, which can be parallelized across multiple GPUs without harming the accuracy. Moreover, introducing a new modality into HyperLearn framework requires only an additional GPU, which makes it scalable to datasets with many modalities. 

\item Extensive experimentation shows that our approach is task-independent, with a potential for deployment in a variety of applications and multimedia collections.



\end{itemize}

\section{Related Work}

The core challenge in multimodal learning revolves around learning representations that can process and relate information from multiple heterogeneous modalities. Most of existing multimodal representation learning methods can be split into two broad categories -- multimodal network embeddings and tensor factorization-based latent representation learning. In this section we reflect on the representative approaches from these two categories. Since, in this work we extend the notion of graph convolution networks for multimodal datasets, we also touch upon some of the existing techniques that aim to deploy deep learning on graphs.
\subsection{Multimodal Network Embedding}
 A common strategy for representation learning is to project different modalities together into a joint feature space. Traditional methods ~\cite{mcauley2012image, roweis2000nonlinear, tenenbaum2000global} focus on generating node embeddings by constructing an affinity graph on the nodes and then finding the leading eigenvectors for representing each node. With the advent of deep learning, neural networks have become a popular
way to construct combined representations. They owe their popularity to the ability to jointly learn high-quality representations in an end-to-end manner. For example, Srivastava and Salakhutdinov proposed an approach for learning higher-level representation for multiple modalities, such as images and texts using Deep Boltzmann Machines (DBM)  \cite{srivastava2012multimodal}. Since then a large number of multimodal representation learning methods based on deep learning have been proposed. Some of these methods  attempt to learn a multimodal network embedding by combining the content and link information ~\cite{huang2018multimodal, li2018deep, tang2015line, zhang2017user, yang2015network, li2017variation, chang2015heterogeneous}. Other set of methods focuses on modeling the correlation between multiple modalities to
learn a shared representation of multimedia items. An example of such coordinated
representation is Deep Canonical Correlation Analysis
(DCCA) that aims to find a non-linear mapping that maximizes the correlation between the mapped vectors from the two modalities \cite{yan2015deep}. Ambiguities often occur while using network embedding methods to learn multimodal relations due to sub-optimal usage of available information. This is mostly because these methods assume relations between items to be pairwise which often leads to loss of information ~\cite{bu2010music, li2013link, arya2018exploiting}.
\subsection{Tensor Factorization Based Latent Representation Learning}
Decoupling a multidimensional tensor into its factor matrices has been proven successful in unraveling latent representations of their components in an unsupervised manner ~\cite{lacroix2018canonical, narita2012tensor, kolda2009tensor}. Most existing approaches aim to embed both entities and relations into a low-dimensional space for tasks such as link prediction \cite{trouillon2016complex}, reasoning in knowledge bases \cite{socher2013reasoning} or multi-label classification problems \cite{maruhashi2018learning}.  Recent methods on social image understanding incorporate user information as the third modality for tag based image retrieval and image-tag refinement problems ~\cite{tang2017tri, tang2019social, tang2015neighborhood}. Even though most of these approaches are suitable for large datasets, one of the main disadvantages of using a factorization based model is the lack of flexibility when scaling to highly multidimensional datasets. Additionally, most of the tensor decomposition methods are based on the optimization with a least squared criterion, which severely lacks robustness to outliers \cite{kim2013robust}.

\par In this work, we first overcome the issues of network embedding methods by using a hypergraph-based learning method. Secondly, we introduce a scalable approach to tensor decomposition for scaling representation learning to many modalities. Finally, we can combine the advantages of rich information from network structure with the unsupervised nature of tensor decomposition in one single end-to-end framework.

\subsection{Geometric Deep Learning on graphs}
Geometric deep learning \cite{bronstein2017geometric} brings the algorithms that can help learn from non-euclidean data like graphs and 3D objects by proposing an ordering of mathematical operators that is different from common convolutional networks. The aim of Geometric Deep Learning is to process signals defined on the vertices of an undirected graph $\mathbb{G}(\textit{V},\textit{E},\textit{W})$, where $\textit{V}$ is the set of vertices, $\textit{E}$ is set of edges, and $\textit{W} \in \mathbb{R}^{|\textit{V}| \times |\textit{V}|}$ is the adjacency matrix. Following \cite{shuman2012emerging, defferrard2016convolutional}, spectral domain convolution of signals $x$ and $y$ defined on the vertices of a graph is formulated as:
\begin{equation}
    x \circledast y = \Phi(\Phi^T x).(\Phi^T y) = \Phi(\mathcal{F}(x).\mathcal{F}(y))
\end{equation}
Here, $\Phi^T x$ corresponds to Graph Fourier Transform and $\mathcal{F}(.)$ represents Fourier Transform; the eigen functions $\Phi$ of the graph laplacian play the role of Fourier modes; the corresponding eigenvalues $\Lambda$ of the graph laplacian are identified as the frequencies of the graph.
Recent applications of graph convolutional networks range from computer graphics \cite{boscaini2016learning} to chemistry \cite{duvenaud2015convolutional}. The spectral graph convolutional neural networks ($GCN$), originally proposed in \cite{bruna2014spectral} and extended in \cite{defferrard2016convolutional} were proven effective in classification of handwritten digits and news texts. A simplification of the \textit{GCN} formulation was proposed in \cite{kipf2016semi} for semi-supervised classification of nodes in a graph. In the computer vision community, \textit{GCN} has been extended to describe shapes in different human poses \cite{masci2015geodesic}, perform action detection in videos \cite{wang2018videos} and for image and 3D shape analysis \cite{monti2017geometric}. However, in the multimedia field there have been considerably less examples of using deep learning on graphs for modeling highly multimodal datasets with \cite{rudinac2017multimodal, arya2018exploiting} as notable exceptions. 

In this paper, we propose an approach that introduces the application of  graph convolutional networks on multimodal datasets. We deploy Multi-Graph Convolution Network (MGCNN) originally proposed by \cite{monti2017geometric} for the matrix completion task using row and column graphs as auxiliary information. It aims at extracting spatial features from a matrix by using information from both the row and column graphs. For a matrix $X \in \mathbb{R}^{N_1 \times N_2}$, MGCNN is given by 
\begin{equation}\label{eq12}
    \widetilde{X} = \sum_{j,j' = 0}^{q} \theta_{jj'} T_j(\mathbb{L}_r) X T_{j'}(\mathbb{L}_c)
\end{equation}
where, $ \Theta = \theta_{jj'}$ is $(q+1)\times(q+1)$ dimensional matrix which represents the coefficients of the filters, $T_j(.)$ denotes the Chebyshev polynomial of degree $j$ and $\mathbb{L}_r$, $\mathbb{L}_c$ are the row and column Graph Laplacians respectively.  Using Equation \ref{eq12} as the convolutional layer of MGCNN, it produces $q$ output channels ($N_1 \times N_2 \times q$) for matrix $X\in \mathbb{R}^{N_1 \times N_2}$ with a single input channel. In this way, one can extract $q$ dimensional features for each item in matrix $X$ by combining information from row and column graphs, which can correspond to e.g. individual modalities.



\section{The Proposed Framework}

\begin{figure}[t!]
  \centering
  \includegraphics[width=\linewidth]{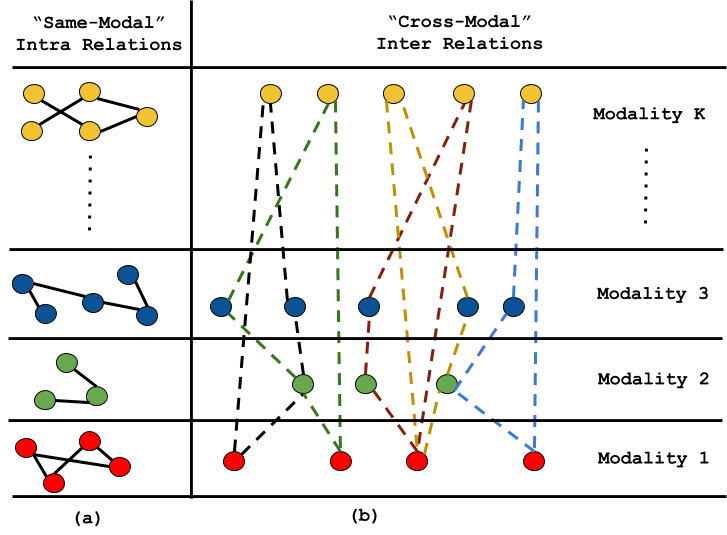}
  \caption{(a) Pair-wise relationship among the items of the same modality in  K-modal data. (b) Complex higher-order heterogeneous relationships between entities of different modalities using a Hypergraph representation.}
  \label{hypergraph}
\end{figure}

\begin{figure*}[t!]
\begin{minipage}[b]{\linewidth}
 \centering
 \centerline{\includegraphics[width=1.0\textwidth]{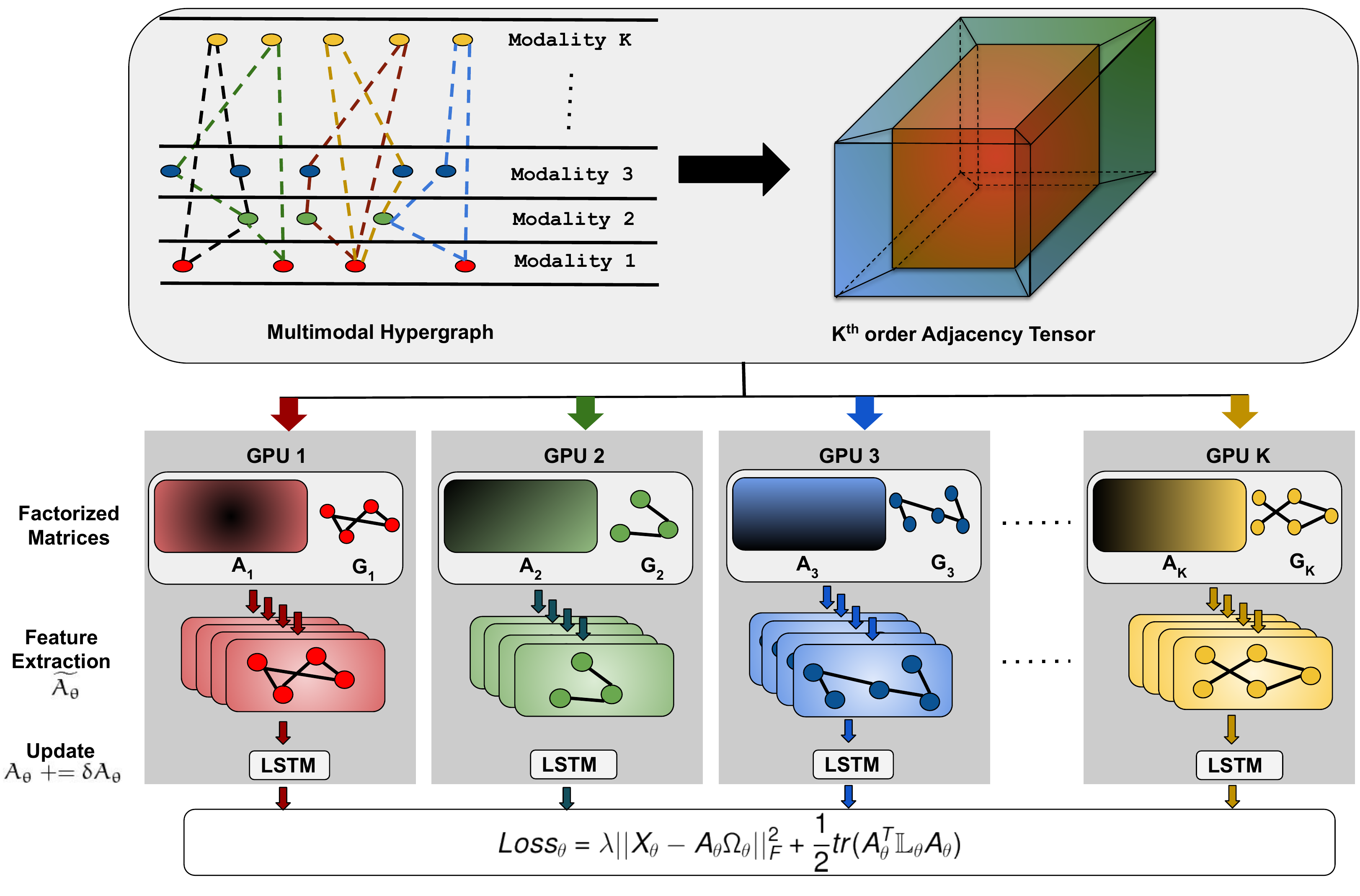}}
 \vspace{-0.8cm}
 \centerline{}\medskip
\end{minipage} 
\caption{ Proposed HyperLearn framework deployed on K modalities with a distributed learning approach}
\label{parallel}
\end{figure*}

In this section, we propose a novel distributed learning framework that can simultaneously exploit both intra and inter-relations in multimodal datasets. We depict these inter-relations on a hypergraph and conjecture that this way of representing higher-order relations reduces any loss of information contained within the multimodal network structure ~\cite{zhou2007learning,li2013link, arya2018exploiting}. Mathematically, a hypergraph is depicted by its adjacency tensor \cite{banerjee2017spectra}. A simple tensor factorization on this adjacency tensor can disentangle modalities into their compact representations. However, this kind of representation lacks information from the intra-relation of items belonging to the same modality. Subsequently, we therefore incorporate intra-relations among entities as auxiliary information to facilitate flow of within-modal relationship information.

\subsection{Notations}
 We use boldface underlined letters, such as $\boldsymbol{\underline{X}}$, to denote tensors and simple upper case letters, such as $U$, to denote matrices. Let $\odot$ represent the "Khatri-Rao" product \cite{khatri1968solutions} defined as
 \begin{equation}
     U \odot V = (U_{ij} \otimes V_{ij})_{ij}
 \end{equation}
 where, $U \in\mathbb{R}^{L \times R}$ and $V  \in\mathbb{R}^{M \times R}$ are arbitrary matrices and $\otimes$ is the Kronecker Product. The resulting matrix $U \odot V$ is an expanded matrix of dimension $LM \times R$ on the columns of $U$ and $V$.
 
\subsection{Representing Cross-Modal Inter-Relations using Hypergraphs} 
Hypergraphs have been proven extremely efficient in depicting higher-order and heterogeneous relations. A hypergraph is the most efficient way to represent complex relationships between a multitude of diverse entities, as it minimizes any loss of available information ~\cite{wolf2016advantages, arya2018exploiting, bu2010music}. Given multimodal data, we construct a unified hypergraph $\mathcal{H}(V,E)$ by building hyperedges ($E$) around each of the individual multimodal items which are represented on a set of nodes ($V$). These hyperedges correspond to the cross-modal relations between items of different modalities as illustrated in Figure \ref{hypergraph}.

\par A more formalised mathematical interpretation of this unified hypergraph is given by its adjacency tensor $\boldsymbol{\underline{X}}$, where the number of components of the tensor is equal to the number of modalities in the hypergraph. Further, each hyperedge corresponds to an entry in the tensor whose value are the weights of the hyperedge. For simplicity, in this work we focus on unweighted hypergraphs.

Thus, a multimodal data with $K$ modalities is depicted on a tensor $\boldsymbol{\underline{X}}\in\mathbb{R}^{N_1 \times N_2 ..... \times N_K}$, where each component $N_\theta$ ($1\leq \theta \leq K$) of this tensor represents one of the $K$ heterogeneous modalities. 
 A single element $\boldsymbol{\underline{x}}$ of $\boldsymbol{\underline{X}}$ is addressed by providing its precise position by a series of indices ${n_1,n_2,..,n_K}$ i.e.
 \begin{equation}
     \boldsymbol{\underline{x}}_{n_1n_2..n_K} \equiv \boldsymbol{\underline{X}}_{n_1n_2..n_K}; \quad 1\leq n_1\leq |N_1|,...,1\leq n_K\leq |N_K| 
 \end{equation}
 
 Further, a hyperedge around a set of nodes can be represented as binary values such that $\boldsymbol{\underline{x}}_{n_1n_2..n_K} =1 $ if the relation $(n_1, n_2,....,n_K)$ is known i.e. if there exists a mutual relation between the $K$ modalities for that instance. For example, in the social network use case, with a possible corresponding image-tag-user associated tensor $ \boldsymbol{\underline{X}} \in \mathbb{R}^{N_1 \times N_2 \times N_3}$, the images ($n_1$) are represented on rows, users ($n_2$) on columns and tags ($n_3$) on tubes. If the $l^{th}$ image uploaded by the $m^{th}$ user is annotated with the {$n^{th}$} tag, then $\boldsymbol{\underline{x}}_{lmn}=1$ and 0 otherwise.

\subsection{Representing Intra-Relations Between the Items of the Same Modality} 
Relationships between items of the same modality are dependent on the nature/properties of the modality. 
For instance, relationships between users in a social network is defined based on their common  interests. To make our framework flexible, each modality ($\theta; 1\leq \theta \leq K$) is represented on a separate graph $\mathbb{\boldsymbol{G_\theta}}$ whose connections can be defined independently. For example: relations among images can be established based on their visual features, for tags it can be calculated based on their co-occurrence and for users it can very well be based on their mutual likes/dislikes. We denote the adjacency matrix of $\mathbb{\boldsymbol{G_\theta}}$ by $ \Lambda _\theta $ where each of its entries $\Lambda^{i,j}_\theta = 1$, if there exists a relation between the $i^{th}$ and $j^{th}$ element and 0 otherwise.
The corresponding normalized graph laplacians ($\mathbb{L}_\theta$) are given by
\begin{equation}
   \mathbb{L}_\theta = D_{\theta}^{\frac{1}{2}}  \Lambda _\theta D_{\theta}^{-\frac{1}{2}}
\end{equation}
where, $D_\theta = diag(\sum_{j\neq i}\Lambda^{i,j}_\theta)$ is known as the degree matrix.

\subsection{Combined Inter-Intra Relational Feature Extraction}

Tensor $\boldsymbol{\underline{X}}$ can be factorized using Candecomp/Parafac(CP) - decomposition \cite{harshman1970foundations} which decomposes a tensor into a sum of outer products of vectors $(a_r^{(\theta)})$.

The CP-decomposition $\widetilde{\boldsymbol{\underline{X}}}$ of $\boldsymbol{\underline{X}}$ is defined as \begin{align}\label{eq4}
        \widetilde{\boldsymbol{X}} &= \sum_{r=1}^R a_r^{(1)}\circ a_r^{(2)}\circ ... \circ a_r^{(K)} \\
        &= \boldsymbol{I}\times_1 A_1 \times_2 A_2 \times_3...\times_K A_K 
\end{align}
where $\circ$ is the outer product and $\times_i$ represents mode-$i$ multiplication (Tensor matrix product). Matrices $A_\theta \in \mathbb{R}^{|N_\theta| \times R}$ are called factor matrices of rank $R$ and $\boldsymbol{I}$ is an $R^{th}$ order identity tensor. Matrices $A_\theta$ are essentially the latent lower dimensional representations for each of the $N_\theta$ components of the tensor and therefore, for each of the $K$ modalities.

\par Subsequently, we introduce an approach that can learn robust representations $A_\theta$ by combining intra relational information. We extract spatial features that merge information from each of the graphs $\mathbb{\boldsymbol{G_\theta}}$ with the latent representation matrices $A_\theta$ using Multi-Graph Convolutional Network (MGCNN) layers given by
\begin{equation}
    \widetilde{A_\theta} = \sum_{j,j' = 0}^{q} \theta_{jj'} T_j(\mathbb{L}_\theta) A_\theta 
\end{equation}
where, the output $\widetilde{A_\theta}\in\mathbb{R}^{|N_\theta| \times R \times q}$ has $q$ output channels. Similar to ~\cite{monti2017geometric}, we use an $LSTM$ to implement the feature diffusion process which essentially iteratively predicts accurate changes $\delta A_\theta$ for the matrix $A_\theta$. Due to its ability to keep long-term internal states, this $LSTM$ architecture is highly efficient in learning complex non-linear diffusion processes.

 \subsection{Loss Function Incorporating Cross-Modality Inter-Relations and Within-Modality Intra Relations}
In standard CP decomposition of a tensor, its factor matrices are approximated by finding a solution to the following equation

\begin{equation}\label{eq7}
     \min_{A_1,..,A_K } ||\boldsymbol{\underline{X}} - (\boldsymbol{I}\times_1 A_1 \times_2 A_2 \times_3...\times_K A_K)||_F^2 
\end{equation}

This equation essentially tries to find low dimensional factor matrices $A_\theta$ such that their combination is as close as possible to the original tensor $\boldsymbol{\underline{X}}$. Further, to add relational information among items within each of these $A_\theta$, we extend the "within-mode" regularization term introduced in \cite{li2009relation} for matrices and \cite{narita2012tensor} for third order tensors to generic $K^{th}$ order tensors. The basic idea is to add a regularization term to Equation \ref{eq7} such that it can force two similar objects in each modality to have similar factors, so that they operate similarly. Thus, the combined loss function is given by:
\begin{equation}\label{eq9}
     \min_{A_1,..,A_K } \frac{1}{2} (tr\sum_{\theta=1}^KA_\theta^T \mathbb{L}_\theta A_\theta)
     + \lambda||\boldsymbol{\underline{X}} - (\boldsymbol{I}\times_1 A_1 \times_2..\times_K A_K) ||^2_F 
\end{equation}
where, $tr(.)$ returns the trace of a matrix. 
In Equation \ref{eq9}, the first term ensures closeness between items of the same modalities and the second term consolidates the relative similarities between items across modalities.
Minimizing Equation \ref{eq9} is a non-convex optimization problem for a set of variables $A_1,..,A_K $. Apart from being an NP-hard problem, computationally it is also expensive to perform even simple operations like element wise product on a $K^{th}$ order tensor. To get a more robust solution, we introduce an alternating method to tensor decomposition similar to ~\cite{kolda2009tensor, kim2013robust}.
 The key insight of such a method is to iteratively solve one of the $K$ components of the tensor while keeping the rest fixed.  We exploit this kind of alternating optimization solution to parallelize our framework across multiple GPUs, by placing each modality on one of them. This creates an independent pipeline for all of the $K$ modalities as shown in Figure~\ref{parallel} which summarizes our distributed learning framework for multimodal datasets. 
 
\subsection{Distributed Training Approach for Learning Latent Representations}
 The separable feature extraction process for each  modality makes our methodology unique and scalable to multiple modalities. These separate pipelines are combined by a joint loss function. Consider solving Equation \ref{eq9} by keeping all other components except $N_{\theta_0}$ as constant. Since, all but one component of the tensor is a variable, unfolding original tensor $\boldsymbol{\underline{X}}$ into a matrix along the $N_{\theta_0}$ component results in matrix $X^{(0)}_\theta$ with dimensions $|N_{\theta_0}| \times |N_1N_2.....N_\theta| $ (where $1 \leq \theta \leq K \quad s.t. \quad \theta \neq \theta_0$). So, the loss function in Equation \ref{eq9} can be rewritten for each of the $K$ components ($N_{\theta}$) as

 \begin{equation}
 Loss_\theta = \lambda ||X_\theta - A_\theta \Omega_\theta||_F^2 + \frac{1}{2} tr(A_\theta^T \mathbb{L}_\theta A_\theta)
 \end{equation}
, where $\Omega_\theta = A_1  \odot A_2 \odot .. \odot A_{\theta-1} \odot A_{\theta+1}.. \odot A_K$ and $\odot$ represents the "Khatri-Rao" product.

\section{Experiments}

\begin{table}[]

\caption{Table showing the total number of intra and inter-relations between items on MovieLens, MIR Flickr and OmniArt datasets. }

\centering
\resizebox{0.45\textwidth}{!}{%
\begin{tabular}{|c||c|c|c|cc}
\cline{1-4}
\textbf{\begin{tabular}[c]{@{}c@{}}Movie\\ Lens\end{tabular}}  & \begin{tabular}[c]{@{}c@{}}$\mathscr{R}$(U)\\ 12,594\end{tabular}   & \begin{tabular}[c]{@{}c@{}}$\mathscr{R}$(M)\\ 28,928\end{tabular} & \begin{tabular}[c]{@{}c@{}}$\mathscr{R}$(U-M)\\ 100,000\end{tabular} &                                                                                 &                                                                                      \\ \cline{1-5}
\textbf{\begin{tabular}[c]{@{}c@{}}MIR\\ Flickr\end{tabular}} & \begin{tabular}[c]{@{}c@{}}$\mathscr{R}$(I)\\ 93,695,167\end{tabular}   & \begin{tabular}[c]{@{}c@{}}$\mathscr{R}$(T)\\ 25,170\end{tabular} & \begin{tabular}[c]{@{}c@{}}$\mathscr{R}$(U)\\ 9,900,716\end{tabular}    & \multicolumn{1}{c|}{\begin{tabular}[c]{@{}c@{}}$\mathscr{R}$(I-T-U)\\ 48,760\end{tabular}} &                                                                                      \\ \hline
\textbf{\begin{tabular}[c]{@{}c@{}}Omni\\ Art\end{tabular}}    & \begin{tabular}[c]{@{}c@{}}$\mathscr{R}$(I)\\ 4,628,009\end{tabular} & \begin{tabular}[c]{@{}c@{}}$\mathscr{R}$(A)\\ 849,482\end{tabular} & \begin{tabular}[c]{@{}c@{}}$\mathscr{R}$(M)\\ 21,178\end{tabular}    & \multicolumn{1}{c|}{\begin{tabular}[c]{@{}c@{}}$\mathscr{R}$($T_f$)\\ 144\end{tabular}}     & \multicolumn{1}{c|}{\begin{tabular}[c]{@{}c@{}}$\mathscr{R}$(I-A-M-$T_f$)\\ 28,399\end{tabular}} \\ \hline
\end{tabular}}
\label{relations}
\end{table}

We start our experimental evaluation showing the performance of our approach on a 2-dimensional standard matrix completion task and then extend it to 3 and 4 dimensional cases. For 2D, 3D and 4D case, we use MovieLens \cite{miller2003movielens}, MIR Flickr \cite{huiskes2008mir} and OmniArt \cite{strezoski2017omniart} datasets respectively. We conjecture that our framework can be generalized to datasets with even more modalities. Table \ref{relations} summarizes the number of inter and intra relations for the three above mentioned cases. Here, $\mathscr{R(.)}$ represents the number of relations. As seen from the table, even relatively small datasets feature a multitude of relations, which makes learning them even more challenging.

\subsection{Task 1: Matrix Completion on Graphs}
\begin{figure}[t]
\centering
\subfloat[Time (in ms) taken for each training iteration \label{fig:main_eff_plot}]{%
  \includegraphics[width=\linewidth]{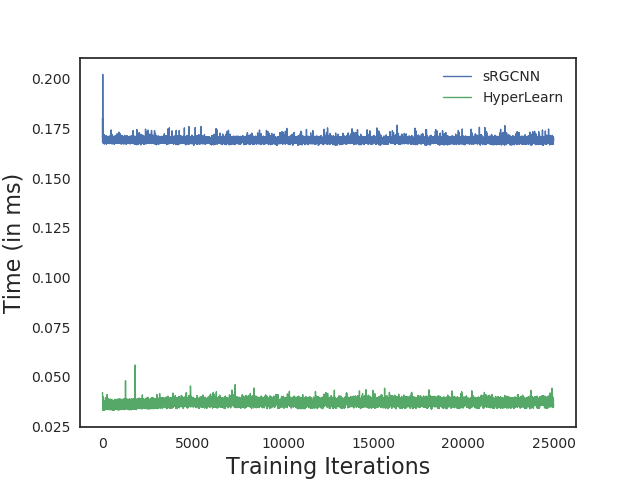}
}\hfill
\subfloat[Convergence rate of RMSE Loss over time \label{fig:inter_plot}]{%
  \includegraphics[width=\linewidth]{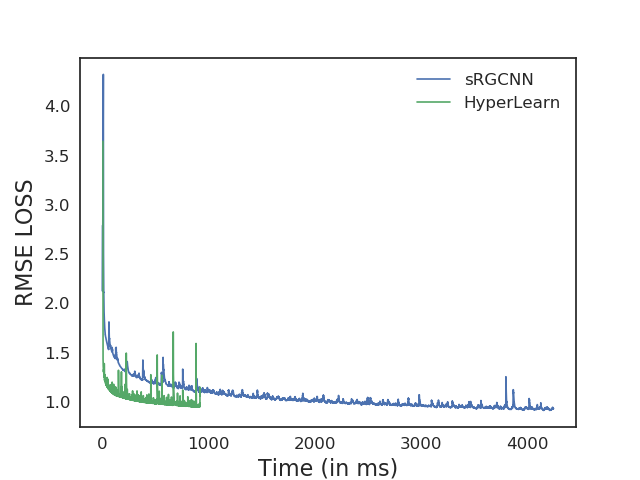}
}%
\caption{Illustration of the convergence rate of HyperLearn against sRGCNN. Our method clearly requires a much lower training time per iteration and also converges much faster than sRGCNN.}
\label{time:epoch:loss}
\end{figure}

\begin{figure*}[t!]
\begin{minipage}[b]{\linewidth}
 \centering
 \centerline{\includegraphics[width=1.0\textwidth]{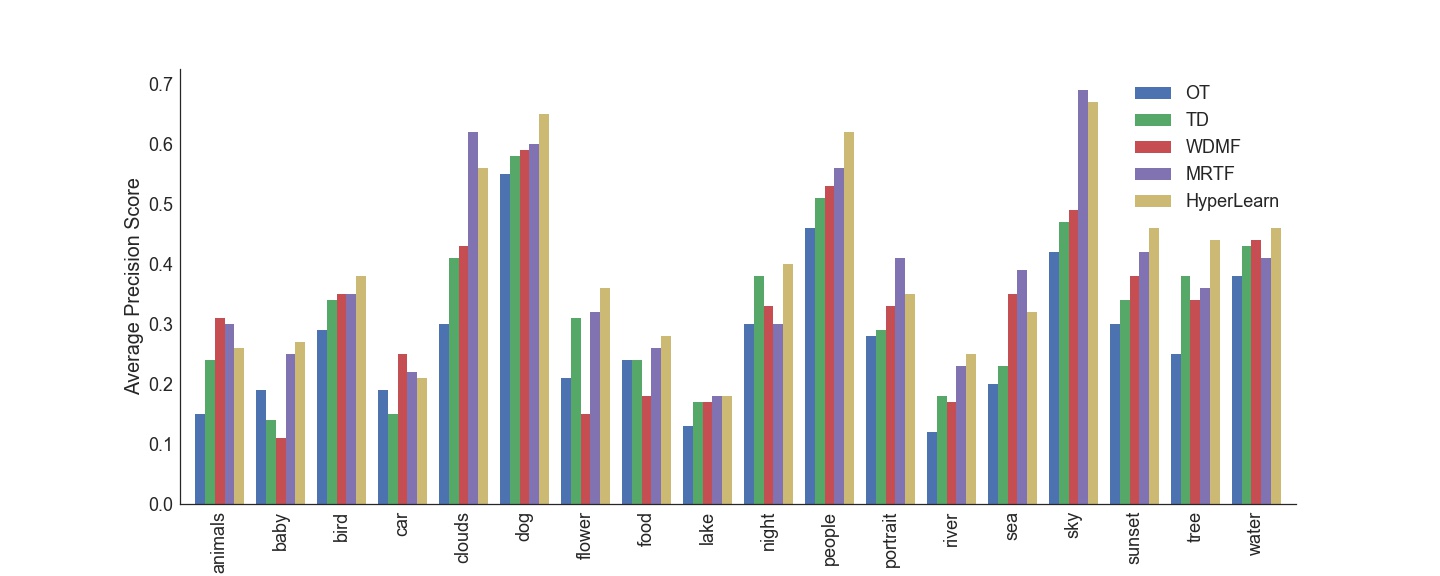}}
 \vspace{-0.8cm}
 \centerline{}\medskip
\end{minipage} 
\caption{ Detailed performance comparison in terms of Average Precision over 18 concepts on the MIRFlickr dataset}
\label{precision}
\end{figure*}

We show the computational advantage of our approach against a matrix completion method that makes use of side information as a baseline. For this, we use the standard MovieLens 100K dataset \cite{miller2003movielens}, which consists of 100,000 ratings on a scale of 0 to 5 corresponding to 943 users ($U$) and for 1,682 movies ($M$). We follow the experimental setup of Monti et.al. \cite{monti2017geometric} for constructing the respective user and movie intra-relation graphs as unweighted 10-nearest neighbor graphs.

We compare the performance of HyperLearn with separable Recurrent Graph Convolutional Networks (sRGCNN) as proposed in \cite{monti2017geometric}.
As can be seen from Figure~\ref{time:epoch:loss}, our approach attains comparable performance to the state of the art alternative, while being much faster. The feature extraction approach alternating between movie and item graphs reduces the time complexity (although not linearly) considerably as can be seen from Figure~\ref{time:epoch:loss}(a) which in turn increases the rate of convergence for the algorithm as depicted in Figure~\ref{time:epoch:loss}(b). However, due to continuous alternating loss calculations, sometimes the back-propagated gradients tend to get biased towards one of the modalities resulting in some higher peaks for HyperLearn in Figure~\ref{time:epoch:loss}(b). 

\subsection{Task 2: Social Image Understanding}

In this experiment, we test the performance of our model on a \nth{3} order multimodal relational dataset. We apply our method to uncover latent image representations by jointly exploring user-provided tagging information, visual features of images and user demographics. We conduct experiments on the social image dataset: MIR Flickr \cite{huiskes2008mir}. The MIR Flickr dataset consists of 25,000 images ($I$) from Flickr posted by 6,386 users ($U$) with over 50,000 user-provided tags ($T$) in total. Some tags are obviously noisy and should be removed. Tags appearing at least 50 times are kept and the remaining ones are removed as in \cite{li2016weakly, tang2017tri}. To include user information, we crawl the groups joined by each user through the Flickr API. Some images have broken links, or are deleted by their users. We remove such images from our dataset which leaves us with 15,662 images, 6,618 users and 315 tags. The dataset also provides manually-created ground truth image annotations at the semantic concept level. For this filtered dataset, there are 18 unique concepts such as animals, bird, sky etc. for the images which we adopt to evaluate the performance. We create an intra-relation graph for images by taking 10-nearest neighbors based on their widely used standard SIFT features. For users, we create edges between them if they joined the same groups and for tags a graph is created based on their co-occurrence.

\par To empirically evaluate the effectiveness of our proposed method, we present the performance of the latent representation of images in classifying them into 18 concepts. We compare our model with the following methods:
\begin{itemize}
\item \textbf{OT}: The user-provided tags from Flickr as baseline.
\item \textbf{TD}: The conventional CANDECOMP/PARAFAC (CP) tensor decomposition \cite{harshman1970foundations}
\item \textbf{WDMF}: Weakly-supervised Deep Matrix Factorization for Social Image Understanding \cite{li2016socializing}
\item \textbf{MRTF}: Multi-correlation Regularized Tensor Factorization approach \cite{sang2011exploiting}
\end{itemize}

\begin{table}[ht]
\caption{Comparison of the training times (in hours) on MIR Flickr dataset}
\begin{tabular}{cc}
\hline \hline
\textbf{Model}      & \textbf{Training Time (in hours)} \\ \hline \hline
WDMF       & 4.2 $\pm$ 0.4            \\ \hline
MRTF       & 2.7 $\pm$ 0.3            \\ \hline
HyperLearn & 1.8 $\pm$ 0.3            \\ \hline 
\end{tabular}
\label{table:time}
\end{table}

These methods, to the best of our knowledge, cannot provide the flexibility of performing distributed training for each modality using multiple GPUs.
We report Average Precision (AP) scores for comparing our HyperLearn approach  against all of these methods. Average Precision (AP) is the standard measure used for multi-label classification benchmarks. It corresponds to the average of the precision at each position where a relevant image appears. Figure~\ref{precision} shows the comparative performance for all the 18 concepts. We also compare HyperLearn with MRTF and WDMF in terms of the training time and report the results in Table~\ref{table:time}. As can be seen from this table, HyperLearn executes faster than MRTF and WDMF while its performance is at par or even better for most of the concepts in the multi-label classification task shown in Figure~\ref{precision}.

Through this experiment we show that - (a) the performance of our approach is at par with the existing methods in understanding social image relationships (b) by introducing a distributed approach we can cut down  training time of the model significantly.

\subsection{Task 3: OmniArt}
In the last experiment, we show the performance of our model in learning relations that go beyond \nth{3} order of connections. For this we require a highly multimodal dataset containing complex relations that are hard to interpret. One such dataset is OmniArt, a large-scale artistic benchmark collection consisting of digitized artworks from museums across the world ~\cite{strezoski2017omniart, strezoski2018omniart}. OmniArt comprises millions of artworks coupled with extensive metadata such as genre, school, material information, creation period and dominant color. This makes the dataset extremely multi-relational and, at the same time, very challenging to perform learning tasks.

For the purpose of comparison with related work, we first perform the artist attribution task in which we attempt to determine the creator of an artwork based on his/her inter-relations with artworks, media (e.g., oil, watercolor, canvas etc.) and creation period (timeframe), along with their intra-relations. To this end we select artworks corresponding to the most common artists in the collection. Considering each of these data streams - artworks $(I)$, artist $(A)$, media $(M)$ and their timeframe ($T_f$) in centuries as a separate modality, we create the inter-relation hypergraph between them. Subsequently, intra-relation graphs are created for each of the 4 modalities in the following way: 
\begin{itemize}
    \item $\mathbb{\boldsymbol{G_I}}$: Based on color palettes similarity
    \item $\mathbb{\boldsymbol{G_A}}$: Based on the schools the artist belongs to
    \item $\mathbb{\boldsymbol{G_M}}$: Based on the co-occurrence in all artworks
    \item $\mathbb{\boldsymbol{G_{T_f}}}$: Based on the style and genre prevalent in that century
\end{itemize}
We take a sub-sample of the OmniArt dataset consisting of 10,000 artworks from 2,776 artists in the time period ranging all the way from \nth{8} to \nth{20} century along with 63 prominent media types. On this sampled dataset, we achieve an accuracy of $61.7\%$ for the artist attribution task. The performance of our model is at par with the benchmark accuracy of $64.5\%$ ~\cite{strezoski2017omniart}. In addition, we conjecture that Hypergraph has an important advantage -- the ability to learn even higher order relations, i.e. \nth{5}, \nth{6} and beyond, something that we intend proving in future work. 

\begin{figure}
\centering
\subfloat[Vincent van Gogh -- Olive Trees with Yellow Sky and Sun, 1889 ]{{\includegraphics[height=2.85cm]{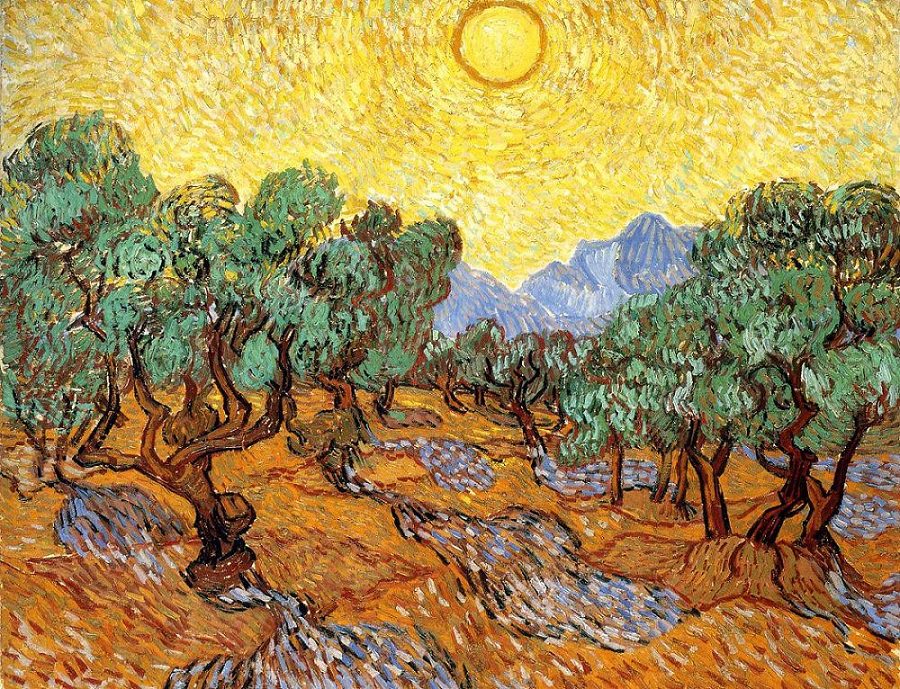} }}%
\quad
\subfloat[Claude Monet -- Marine View with a Sunset, 1875]{{\includegraphics[height=2.85cm]{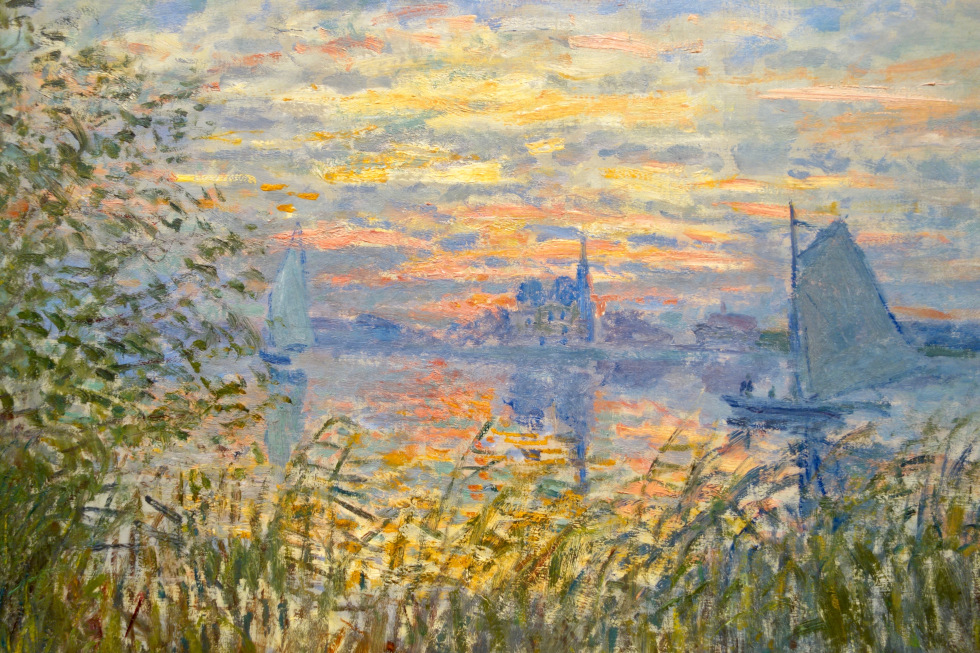} }}%
\caption{Van Gogh (a) and Monet (b) have many stylistic similarities, but their materialization is different. Capturing their similarities, differences and influences requires the ability to model higher-order relations.}%
\label{fig:vangogh:mone}%
\end{figure}

In the particular case of OmniArt, such higher-order relations would include information about e.g., artist, school, timeframe, medium, dominant colour use, semantics and (implicit) social network. For example, Figure~\ref{fig:vangogh:mone} shows the well-known ``Olive Trees with Yellow Sky and Sun'' painted by Vincent van Gogh in 1889 and Claude Monet's masterpiece ``Marine View with a Sunset'' from 1875. As nicely portrayed by these two examples, while the two artists exhibit many stylistic similarities, sharing motives and a time period, their materialization is very different. Influenced by Monet, Van Gogh changed both his colour palette and coarseness of brushstrokes, so technically, his work became closer to the French Impressionism. Detecting ``tipping points'' in the artist's opus would require multimedia representations capable of capturing information about e.g. colour, texture and semantic concepts depicted in the paintings, but also information about school, social network, relevant locations and timeframe and historical context. We believe that our proposed framework is a significant and brave step forward in ultimately deploying multimedia analysis for solving such complex tasks. 

\section{Conclusion and Future Work}
In this paper we propose HyperLearn, a hypergraph-based framework for learning complex higher-order relationships in multimedia datasets. The proposed distributed training approach makes this framework scalable to many modalities. We demonstrate benefits of our approach with regards to both performance and computational time through extensive experimentation on MovieLens and MIRFlickr datasets with 2 and 3 modalities respectively.
To show the flexibility of HyperLearn in encoding a larger number modalities, we perform experiments on \nth{4} order relations from the OmniArt dataset. In conclusion, on the examples of very different datasets, domains and use cases, we demonstrate that HyperLearn can be extremely useful in learning representations that can capture complex higher order relations within and across multiple modalities. For future work we plan to test the approach on even higher number of heterogeneous modalities and further extend this approach to much larger datasets by solving sub-tensors derived from slicing hypergraph into multiple smaller hypergraphs.

\bibliographystyle{ACM-Reference-Format}
\bibliography{arxiv_version}
\end{document}